\documentclass[twoside]{article}

\usepackage[accepted]{aistats2023}
\usepackage[round]{natbib}

\usepackage{amsmath}
\usepackage{amssymb}
\usepackage{amsthm}
\usepackage{mathtools}
\usepackage{multirow}
\usepackage{xcolor}
\usepackage{algorithm}
\usepackage{comment}
\usepackage{bbold}
\usepackage{threeparttable,booktabs}
\usepackage[noend]{algpseudocode}
\usepackage{graphicx}
\usepackage{caption}
\usepackage{subcaption}
\usepackage{enumitem}

\renewcommand{\vec}[1]{\boldsymbol{#1}}
\newcommand{\given}{\, | \,}

\newcommand{\cat}{\operatorname{Cat}}

\newcommand*{\defeq}{\mathrel{\vcenter{\baselineskip0.5ex \lineskiplimit0pt
			\hbox{\footnotesize.}\hbox{\footnotesize.}}}%
	=}

\newcommand{\cX}{\mathcal{X}}
\newcommand{\cY}{\mathcal{Y}}

\newcommand{\Prob}{P}
\newcommand{\prob}{\vec{p}}

\newcommand{\argmax}{\operatorname*{argmax}}

\newtheorem{definition}{Definition}

\begin{document}
\runningauthor{Mortier, Bengs, H\"ullermeier, Luca, Waegeman}

%

%

\twocolumn[

\aistatstitle{On the Calibration of Probabilistic Classifier Sets}

\aistatsauthor{
	Thomas Mortier\\
	\And
	Viktor Bengs\\
	\And
	Eyke H\"ullermeier\\
	\And
	Stijn Luca\\
	\And
	Willem Waegeman\\
}

\aistatsaddress{
	KERMIT\\
	Ghent University
	\And
	Institute of Informatics\\
	LMU Munich
	\And
	Institute of Informatics\\
	LMU Munich
	\And
	KERMIT\\
	Ghent University
	\And
	KERMIT\\
	Ghent University
} ]

\begin{abstract}
Multi-class classification methods that produce sets of probabilistic classifiers, such as ensemble learning methods, are able to model aleatoric and epistemic uncertainty. Aleatoric uncertainty is then typically quantified via the Bayes error, and epistemic uncertainty via the size of the set. In this paper, we extend the notion of calibration, which is commonly used to evaluate the validity of the aleatoric uncertainty representation of a single probabilistic classifier, to assess the validity of an epistemic uncertainty representation obtained by sets of probabilistic classifiers.  
Broadly speaking, we call a set of probabilistic classifiers calibrated if one can find a calibrated convex combination of these classifiers. To evaluate this notion of calibration, we propose a novel nonparametric calibration test that generalizes an existing test for single probabilistic classifiers to the case of sets of probabilistic classifiers. Making use of this test, we empirically show that ensembles of deep neural networks are often not well calibrated.
\end{abstract}

\section{INTRODUCTION}

There is a general consensus that trustworthy machine learning systems should not only return accurate predictions, but also a credible representation of their uncertainty. In this regard, two inherently different sources of uncertainty are often distinguished, referred to as \emph{aleatoric} and \emph{epistemic} \citep{hora_aa96}, and various methods that quantify these types of uncertainty have been proposed \citep{mpub272,kend_wu17,mpub440}. In multi-class classification problems, which will be the setting of interest in this paper, such methods typically focus on the uncertainty in the outcome $y \in \mathcal{Y}$ given a query instance $\vec{x} \in \cX$ for which a prediction is sought. 

\paragraph{Background.} Consider a standard multi-class classification setting with instance space $\cX$ and label space $\cY = \{ 1 , \ldots ,K \}$. We assume that the data is i.i.d.\ according to an underlying joint probability measure $\Prob$ on $\cX \times \cY$. Correspondingly, each instance $\vec{x} \in \cX$ is associated with a conditional distribution $\prob(\vec{x})$, where the component $ p_k(\vec{x}) = p( Y=k \given \vec{x})$ is the probability to observe class $k$ as an outcome given $\vec{x}$. 
A probabilistic classifier $\hat{\vec{p}}$  can then be defined as an estimator of $\prob(\vec{x})$. Usually, $\hat{\vec{p}}$ is learned from a hypothesis space $\mathcal{H}$, which is a subset of all possible mappings from $\mathcal{X}$ to $\Delta_K$, where $\Delta_K$ denotes the $(K-1)$-simplex
 \begin{equation}
\Delta_K \defeq	  \big\{ \vec{\theta} = (\theta_1, \ldots , \theta_K) \in [0,1]^K \given \| \vec{\theta} \|_{1} = 1 \big\}  
\end{equation}
of probability vectors $\vec{\theta}$. Each of these vectors identifies a categorical distribution $\cat(\vec{\theta})$. 
Let us remark that we will utilize bold lowercase letters for vectors and vector functions, e.g., $\vec{\theta}$ and $\hat{\vec{p}}(\vec{x})$, whereas a normal font will depict the components of these vectors, e.g., $\theta_k$ and $\hat{p}_k(\vec{x})$. 

In this setting, aleatoric uncertainty is usually defined as uncertainty that cannot be reduced by collecting more training samples, as it originates from wrong class annotations or a lack of informative features. Therefore, the ``ground-truth'' is the above-mentioned conditional probability distribution $\prob(\vec{x})$, and the Bayes error of this distribution tells us whether the aleatoric uncertainty is high. Even with perfect knowledge about the underlying data-generating process, the outcome cannot be predicted with certainty. 
However, the learner does not know $\prob(\vec{x})$ in practice. Instead, it uses an estimate $\hat{\prob}(\vec{x})$ based on the training data as a surrogate. In essence, epistemic uncertainty refers to the uncertainty about the true $\prob$, or the ``gap''  between $\prob$ and $\hat{\prob}$.
Methods that represent this (second-order) uncertainty typically do not return a single $\hat{\prob}( \vec{x})$, but either a distribution over $\hat{\prob}(\vec{x})$, as in Bayesian methods, or sets containing different $\hat{\prob}( \vec{x})$, as in ensemble methods. 

The validity of an aleatoric uncertainty representation is typically evaluated by statistically testing whether a fitted model $\hat{\prob}( \vec{x})$ is calibrated, see e.g.\ \citep{Hosmer2003,Vaicenavicius2019,Widmann2019}. However, to be very precise, calibration tests do not test whether a model $\hat{\prob}(\vec{x})$ corresponds to the ground-truth $\prob(\vec{x})$, because calibration is only a necessary condition. Likewise, since the true $\prob(\vec{x})$ is in practice never observed, the evaluation of epistemic uncertainty representations is also far from trivial. For lack of an objective ground-truth, most methods assess epistemic uncertainty representations in an indirect manner, using downstream tasks such as out-of-distribution detection \citep{Ovadia2019}, robustness to adversarial attacks \citep{KopetzkiCZGG21} or active learning \citep{NguyenSH22}. However, all those tasks are characterized by a scenario where training and test data are not identically distributed. Therefore, several recent studies have raised concerns about the usefulness of such tasks w.r.t.\ epistemic uncertainty evaluation \citep{Abe2022,Dewolf2021,Bengs2022}. 


\paragraph{Contribution.} In this work, we present a statistical approach to test the validity of epistemic uncertainty representations of methods that use sets of distributions for prediction purposes. This includes various types of ensemble methods, such as random forests \citep{mpub416}, deep ensembles \citep{Lakshminarayanan17,Rahaman21UQDE,Schulz22LPDL}, dropout networks \citep{Gal2016}, stacking methods \citep{Ting1999} and linear opinion pooling \citep{Hora04LP,Ranjan10LP,Lichtendahl13LP}.
With ensemble methods and a predefined hypothesis space $\mathcal{H}$ of probabilistic classifiers, one can assume that in total $M$ different probabilistic models are fitted to the training data, denoted as $\mathcal{P} = \{\hat{\vec{p}}^{(1)},\ldots,\hat{\vec{p}}^{(M)}\}\subseteq \mathcal{H}$, where $\hat{\vec{p}}^{(m)}$ represents the $m$-th model. 

Inspired by the literature on linear opinion pools and stacking, we are interested in the convex set that contains all convex combinations of these $M$ models: 
\begin{equation}
\arraycolsep=0pt
\label{eq:convexset}
S(\mathcal{P}) = \left\{ \hat{\vec{p}}_{\vec{\lambda}}  \, \Big|\, \hat{\vec{p}}_{\vec{\lambda}}(\vec{x}) = \sum\limits_{m=1}^M \lambda_m \hat{\vec{p}}^{(m)} (\vec{x}), \vec{\lambda} \in \Delta_M \right\} 
\end{equation}
with $\Delta_M \defeq	  \big\{ \vec{\lambda} = (\lambda_1, \ldots , \lambda_M) \in [0,1]^M \given \| \vec{\lambda} \|_{1} = 1 \big\}$. 
Two properties are of crucial importance for these convex sets. On the one hand, the \emph{size} of the convex set quantifies the degree of epistemic uncertainty, so this size should be as small as possible. On the other hand, the convex set should be a \emph{valid} representation, i.e., it should contain the true $\prob$. To measure the size of the convex set, which is typically done in a pointwise manner, various measures are used in the literature, such as the mutual information \citep{Malinin20Ensembles} or the generalized Hartley measure \citep{Hullermeier22CU}. Conversely, to evaluate the validity, no method exists today. 


Therefore, we introduce in this paper the first calibration test to evaluate the validity of set-based epistemic uncertainty representations, starting from existing calibration tests for aleatoric uncertainty representations. Informally, we call a set-based epistemic uncertainty representation calibrated if there exists a calibrated convex combination in the set of Eq.~(\ref{eq:convexset}). This testing problem gives rise to a challenging scenario of multiple hypothesis testing, since all members in the corresponding convex set need to be tested (in the worst case). We propose a novel test based on resampling and analyze empirically the Type I and Type II error for various calibration measures and conditions. Lastly, we apply our newly-developed test to analyze whether deep ensembles and dropout networks represent epistemic uncertainty in a correct manner. 

\section{ALEATORIC UNCERTAINTY EVALUATION}
\label{sec:aleatoric}

In this section we formally review existing calibration tests for multi-class classification problems. With such tests, aleatoric uncertainty representations of classifiers can be evaluated in a natural and direct way. In the following section, these tests will be extended for epistemic uncertainty representations in the form of probabilistic classifier sets. This section can hence be interpreted as a discussion of closely-related work, but it also intends to introduce the mathematical concepts needed further on.   

\subsection{Different Notions Of Calibration}

We start by formally defining confidence calibration \citep{Guo2017}. This notion of calibration is by far the most often used in literature \citep{Menezes}. Let's assume that we have fitted a probabilistic classifier that estimates $\prob(\vec{x})$, leading to the estimate $\hat{\vec{p}}(\vec{x})$. Furthermore, let $c(\vec{x}) = \max_k \hat{p}_{k}(\vec{x})$ be the mode of the estimated conditional class distribution for instance $\vec{x}$ (or confidence score) and let $f(\vec{x}) = \argmax_k \hat{p}_{k}(\vec{x})$ be the corresponding prediction.  

\begin{definition}
A multi-class classifier is \emph{confidence calibrated} if it holds that
$$\mathbb{P}_{(\vec{X},Y) \sim \Prob} \, \left(Y = f(\vec{X}) \,|\, c(\vec{X}) = s \right) = s \,.$$
\end{definition}

This is a rather weak notion of calibration, as only the mode of the conditional distribution needs to be calibrated. A stronger form of calibration is classwise calibration \citep{Zadrozny2001}. 

\begin{definition}
A multi-class classifier is \emph{classwise calibrated} if for all $k \in \{1,...,K\}$ it holds that
$$\mathbb{P}_{(\vec{X},Y) \sim \Prob} \, \left(Y = k \,|\, \hat{p}_{k}(\vec{X}) = s \right) = s \,.$$
\end{definition}

A few authors define an even stronger notion of calibration, sometimes referred to as calibration in the strong sense \citep{Widmann2019,Menezes}.  

\begin{definition} \label{defi_strong_calibration}
A multi-class classifier is \emph{calibrated in the strong sense} if for all $k \in \{1,...,K\}$ it holds that
$$\mathbb{P}_{(\vec{X},Y) \sim \Prob} \, \left(Y = k \,|\, \hat{\vec{p}}(\vec{X}) = \vec{s} \right) = s_k \,,$$
with $s_k$ the $k$-th component of $\vec{s}$. 
\end{definition}

\subsection{Calibration Tests} \label{subsec_calib_test}
All three definitions of calibration assume that one knows the true underlying distribution $\Prob$. However, in practice, this distribution is unknown, so one needs to replace the expectation by a finite-sample estimate. In addition, a calibration measure that quantifies the discrepancy between observed and expected frequencies is needed. In a final step, this calibration measure can be used to construct a statistical test that decides whether a multi-class classifier is calibrated or not. In the literature, three different types of tests have been developed. The Hosmer-Lemeshow test is a specific type of chi-squared test that is commonly used as a goodness-of-fit test for logistic and multinomial regression models in statistics~\citep{Hosmer2003,Fagerland2008}. The resampling-based test of \citet{Vaicenavicius2019} and the kernel-based  test of \citet{Widmann2019} have been proposed more recently in the machine learning literature. The Hosmer-Lemeshow test can only be used to assess classwise calibration, whereas the other tests are more general, because they can be used in combination with a wide range of calibration measures. All tests analyze the null hypothesis $H_0: \hat{\vec{p}} \mbox{ is calibrated}$ versus the alternative $H_1: \hat{\vec{p}} \mbox{ is not calibrated}$, where the notion of calibration (i.e., confidence, classwise or strong) depends on the underlying test.   
	
\textbf{\citet{Hosmer2003}} originally proposed the Hosmer-Lemeshow (HL) test for binary logistic regression, and later \citet{Fagerland2008} extended it to the more general class of multinomial probabilistic models. Let $\mathcal{D}_{val} = \{ (\vec{x}_1 , y_1),...,(\vec{x}_N,y_N)\}$ be a validation set of size $N$, i.i.d.\ according to $\Prob$. We will use the shorthand notation $\hat{p}_{ik}$ for the estimated probability $\hat{p}_k(\vec{x}_i)$. Let $c_i = \max_k \hat{p}_{ik}$ be the highest probability for instance $\vec{x}_i$ and let $\hat{y}_i = \argmax_k \hat{p}_{ik}$ be the predicted label. For every label $y_i$, let us consider a $K$-dimensional vector $\vec{y}_i$ that defines the one-hot encoding of $y_i$, i.e., $y_{ik} = 1$ iff $y_i = k$. Similarly as for the true label, we transform $\hat{y}_i$ to a $K$-dimensional vector $\hat{\vec{y}}_i$ that defines the one-hot encoding of $\hat{y}_i$, i.e., $\hat{y}_{ik} = 1$ iff $\hat{y}_i = k$. 

For class $k$, the probabilities $\hat{p}_{1k},...,\hat{p}_{Nk}$ are sorted, and the corresponding instances are placed into equal-frequency bins $\mathcal{B}_{1k},...,\mathcal{B}_{Bk}$ with $B$ the number of bins. Furthermore, the following test statistic is considered:
\begin{equation}
\label{eq:HL}
HL_{cwise}(\hat{\vec{p}},\mathcal{D}_{val}) = \sum_{k=1}^K \sum_{j=1}^B \frac{(o(\mathcal{B}_{jk}) - \mathrm{p}(\mathcal{B}_{jk}))^2}{\mathrm{p}(\mathcal{B}_{jk})} \,,
\end{equation}
\begin{align}
\label{eq:obsexp}
\mbox{with } \qquad  o(\mathcal{B}_{jk}) &= \frac{1}{|\mathcal{B}_{jk}|} \sum_{i: \hat{p}_{ik} \in \mathcal{B}_{jk}}  y_{ik}\,, \nonumber \\
\mathrm{p}(\mathcal{B}_{jk}) &= \frac{1}{|\mathcal{B}_{jk}|} \sum_{i: \hat{p}_{ik} \in \mathcal{B}_{jk}} \hat{p}_{ik} \,,
\end{align}
where $o(\mathcal{B}_{jk})$ and $\mathrm{p}(\mathcal{B}_{jk})$ denote the observed and expected probabilities in bin $\mathcal{B}_{jk}$, respectively. 
\citet{Fagerland2008} argued that (\ref{eq:HL}) follows approximately a chi-squared distribution with $(K-1)(B-2)$ degrees of freedom, and they derived p-values using this assumption. Let us remark that, besides the HL test, many other alternative goodness-of-fit tests exist in statistics, but these usually do not assess model calibration in a direct manner. 

\textbf{\citet{Vaicenavicius2019}} developed a more general (nonparametric) test that can be adopted to assess confidence calibration, classwise calibration and calibration in the strong sense, depending on the calibration measure that is used. This test constructs a bootstrap distribution of any calibration measure empirically, under the null hypothesis that a classifier is calibrated, by resampling new labels multiple times from the probabilistic model that is under assessment. 
After constructing the distribution of the calibration measure under the null hypothesis, the test verifies, for the observed labels, how likely the calibration measure is under the assumption of a calibrated model.  

The test of \citet{Vaicenavicius2019} is often used with the expected calibration error (ECE) as calibration measure. This measure has been originally introduced for binary problems as a way to evaluate reliability diagrams in a quantitative manner. The classwise extension of this measure in fact looks very similar to the Hosmer-Lemeshow test statistic, so it is used to assess classwise calibration (Def. 2): 
\begin{equation}
\label{eq:ECEcwise}
ECE_{cwise}(\hat{\vec{p}},\mathcal{D}_{val}) = \frac{1}{K}\sum_{k=1}^K \sum_{j=1}^B \frac{|\mathcal{B}_{jk}|}{N} |o(\mathcal{B}_{jk}) - \mathrm{p}(\mathcal{B}_{jk})| \,,
\end{equation}
with $o(\mathcal{B}_{jk})$ and $\mathrm{p}(\mathcal{B}_{jk})$ as in (\ref{eq:obsexp}). However, for the expected calibration error, the $[0,1]$-interval is often subdivided in intervals of equal length instead of equal frequency, i.e., $\mathcal{B}_{jk} := \{i: \frac{j-1}{B} \leq \hat{p}_{ik} < \frac{j}{B}\}$. 

For the confidence-based extension of expected calibration error, which is used to assess confidence calibration (Def. 1), binning only has to be done once (so, not for every class $k$). Let us divide the unit interval into $B$ subintervals of equal length. With the $j$-th subinterval we associate the bin $\mathcal{B}_j := \{i: \frac{j-1}{B} \leq c_i < \frac{j}{B}\}$\footnote{the last bins $\mathcal{B}_{Bk}$ and $\mathcal{B}_{B}$ are right-closed.}. Then, the calibration measure becomes:
\begin{equation}
\label{eq:ECEconf}
ECE_{conf}(\hat{\vec{p}},\mathcal{D}_{val}) = \sum_{j=1}^B \frac{|\mathcal{B}_j|}{N} |\mathrm{Acc}(\mathcal{B}_j) - \mathrm{c}(\mathcal{B}_j)| \,,
\end{equation}
\begin{align*}
  \mbox{with} \quad \mathrm{Acc}(\mathcal{B}_j) &= \frac{1}{|\mathcal{B}_j|} \sum_{i: c_i \in \mathcal{B}_j} \sum_{k=1}^K y_{ik} \hat{y}_{ik}\,, \\ 
  \mathrm{c}(\mathcal{B}_j) &= \frac{1}{|\mathcal{B}_j|} \sum_{i: c_i \in \mathcal{B}_j} c_i \,.
\end{align*}



\textbf{\citet{Widmann2019}} proposed a class of measures derived from matrix-valued kernel functions, the so-called squared kernel calibration error (SKCE), to test calibration in the strong sense, without the need for binning.
For example, if one analyses all pairs of instances, while using a general matrix-valued kernel $\Gamma: \Delta_K \times \Delta_K \rightarrow \mathbb{R}^{K \times K}$, then the measure becomes:
\begin{align}
\label{eq:SKCEuq}
\widehat{SKCE}_{uq}(\hat{\vec{p}}, \mathcal{D}_{val}) &= {N \choose 2}^{-1}\nonumber \\ 
& \times \sum_{i=1}^{N-1} \sum_{j=i+1}^N \sum_{s=1}^K \sum_{t=1}^K (\hat{p}_{is} - y_{is}) \nonumber \\
& \times (\hat{p}_{jt} - y_{jt}) \Gamma_{st}(\hat{\vec{p}}(\vec{x}_i),\hat{\vec{p}}(\vec{x}_j)) \,.
\end{align}
However, it is clear that calculating the above measure is computationally expensive when the number of instances $N$ increases. For that reason, a similar estimator was proposed by the same authors:
\begin{align}
\label{eq:SKCEul}
& \widehat{SKCE}_{ul}(\hat{\vec{p}}, \mathcal{D}_{val}) = \frac{1}{\lfloor N/2 \rfloor} \nonumber \\
& \times \sum_{i=1}^{\lfloor N/2 \rfloor}\sum_{s,t=1}^K  (\hat{p}_{(2i-1)s} - y_{(2i-1)s})(\hat{p}_{2it} - y_{2it}) \nonumber \\
& \times \Gamma_{st}(\hat{\vec{p}}(\vec{x}_{2i-1}),\hat{\vec{p}}(\vec{x}_{2i})) \,,
\end{align}
which is linear in $N$. \citet{Widmann2019} also consider other calibration measures, and they derive bounds and approximations on the p-value for the null hypothesis $H_0$ that the model is calibrated. These tests require a lot of space to explain, so we refer to the original paper for a further discussion.

\section{EPISTEMIC UNCERTAINTY EVALUATION}
In this section we develop calibration tests for epistemic uncertainty representations of type $S(\mathcal{P})$, as defined in Eq.~(\ref{eq:convexset}), starting from the methodology that was reviewed in the previous section. 
In the literature, this type of set is sometimes called a credal set, which is typically assumed to be convex and closed~\citep{Walley1991,Corani2014,Yang2014}. Furthermore, from a statistical point of view, there is also a close connection to linear opinion pooling, where the goal is to aggregate probability expert forecasts in an appropriate way, e.g. keeping in mind a proper scoring rule~\citep{Hora04LP,Ranjan10LP,Lichtendahl13LP}. 
In this work, $S(\mathcal{P})$ represents epistemic uncertainty, i.e., due to a limited training dataset, one cannot estimate the ground-truth probability distribution precisely, but this distribution should be contained in the set. An interpretation of that kind allows us to present natural extensions of confidence calibration, classwise calibration, and calibration in the strong sense for probabilistic classifier sets.

\begin{definition}
\label{def:pcs}
A probabilistic classifier set $S(\mathcal{P})$ is confidence calibrated (cf.\ classwise calibrated or calibrated in the strong sense) if there exists a probabilistic model $\hat{\vec{p}}_{\vec{\lambda}}(\vec{x}) \in S(\mathcal{P})$ that is confidence calibrated (cf.\ classwise calibrated or calibrated in the strong sense).
%
\end{definition}

In what follows, we develop a statistical test to verify whether a probabilistic classifier set is calibrated according to Def.~\ref{def:pcs}, which translates to the following  hypotheses:
\begin{equation}
  \label{def_test_single_prob}
H_0: S(\mathcal{P}) \mbox{ is calibrated},\,H_1: \neg H_0
\end{equation}
%
		%
		%
		%
%
The above set of hypotheses can also be written as: 
\begin{equation}
  \label{def_test_ensemble_prob}
H_0: \exists \vec{\lambda}\in \Delta_M \mbox{ s.t.\ } \hat{\vec{p}}_{\vec{\lambda}} \mbox{ is calibrated},\, H_1: \neg H_0
\end{equation}
Furthermore, in what follows, we will use the generic term ``calibrated'' or ``calibration'' to denote any notion of calibration.
%
	%
		%
		%
		%
	%
%
While the test problem in Section \ref{subsec_calib_test} focuses on the question whether a \emph{single} fixed probabilistic classifier $\hat{\vec{p}}$ is calibrated, the test problem in \eqref{def_test_single_prob} or \eqref{def_test_ensemble_prob} asks for the existence of a calibrated convex combination of the finite set of probabilistic classifiers. Thus, the latter is essentially a multiple comparison problem, as we simultaneously test for all $\vec{\lambda}$ the hypotheses 
\begin{equation}
H_{0,\vec{\lambda}}: \hat{\vec{p}}_{\vec{\lambda}} \mbox{ is calibrated},\, H_{1,\vec{\lambda}}:\neg H_{0,\vec{\lambda}}
\end{equation}
%
%
%
%
%
%
Since the number of possible convex combinations is infinite, one cannot resort to standard ways for addressing multiple hypothesis testing problems, such as Bonferroni-correction or the Holm-Bonferroni method. In what follows, we will adopt the extreme value approach for high-dimensional testing \citep{Dickhaus2015}. 
All calibration measures considered in Section~2 are in essence decreasing functions of the \emph{degree of calibration} of a probabilistic model, so one can simply search for the minimum of any calibration measure of interest over $\vec{\lambda} \in \Delta_M$. Then, the multiple hypothesis testing problem in (\ref{def_test_ensemble_prob}) can be addressed by considering the distribution of the minimum under the null hypothesis. 

This principle leads to a nonparametric test that is in fact a direct generalization of the test of \citet{Vaicenavicius2019}. The pseudocode of this procedure is given in Alg.~\ref{alg:gcr}. Starting from a set $\mathcal{P}$ containing $M$ fitted probabilistic models, a validation dataset $\mathcal{D}_{val} = \{ (\vec{x}_1 , y_1),...,(\vec{x}_N,y_N)\}$, and a general calibration measure $g$, such as $HL_{cwise}$ or $ECE_{conf}$, the method first constructs the distribution of the calibration measure under the null hypothesis. To this end, it performs in total $D$ runs with $D$ a hyperparameter that trades off runtime versus p-value precision. In each run, a bootstrap sample of size $N$ is sampled from the original validation dataset (line 2). Then, one $\vec{\lambda} \in \Delta_M$ is selected at random and the corresponding element of the convex set $S(\mathcal{P})$ is chosen as the ground-truth probability distribution, since we are working under the assumption that the null hypothesis is true (line 3). We assume that every element of $S(\mathcal{P})$ is equally likely to correspond to the ground-truth distribution $P$, thus a uniform sample is drawn. Subsequently, for every instance in the bootstrap replicate, a label is randomly drawn from the selected probability distribution (line 4). In the last step of every run, the calibration measure is computed for the generated artificial labels and the ground-truth probability distribution (line 5). After $D$ runs, we assume that a good estimate of the distribution of the calibration measure is obtained, under the assumption that the null hypothesis is true. Given a controlled Type I error rate $\alpha$, the $(1-\alpha)$-quantile of this distribution is computed (line 6). This quantile defines the maximum allowed value of the calibration measure to not reject the null hypothesis. Subsequently, the minimum of the calibration measure is computed for all members of the convex set, using now the original validation dataset (line 7). The null hypothesis is rejected if this minimum exceeds the threshold that was found under the empirical null distribution (line 8). 

\begin{algorithm}[t!]
\begin{small}
\caption{Calibration Test for Probabilistic Classifier Sets -- \textbf{input:}  $\mathcal{P},\mathcal{D}_{\text{val}},g,\alpha,D$}
\begin{algorithmic}[1] 
\For{$d=1,\ldots,D$}
    \State $\mathcal{D}_{d} \leftarrow $ extract bootstrap sample of size $N$ from $\mathcal{D}_{\text{val}}$ (only features are further used)
    \State Sample uniformly $\boldsymbol{\lambda}_{0,d} \in \Delta_{M}$ 
	\State For all $\vec{x}_i \in \mathcal{D}_d$, sample $y_i$ from  $Cat(\hat{\vec{p}}_{\vec{\lambda}_{0,d}}(\vec{x}_i))$ 
    \State $t_{0d} \leftarrow g(\hat{\vec{p}}_{\vec{\lambda}_{0,d}},\mathcal{D}_{d})$ \Comment{$g$ represents an arbitrary calibration measure}
\EndFor
\State $q_{1-\alpha} \leftarrow $ compute $(1-\alpha)$-quantile of empirical distribution $\{t_{0,1},\ldots,t_{0,D}\}$
\State $t \leftarrow \min_{\boldsymbol{\lambda} \in \Delta_M} g(\hat{\vec{p}}_{\boldsymbol{\lambda}},\mathcal{D}_{\text{val}})$ \Comment{Use an appropriate optimization algorithm for $g$}
\State reject $H_0$ if $t>q_{1-\alpha}$ and don't reject $H_0$ otherwise
\end{algorithmic}
\label{alg:gcr}
\end{small}
\end{algorithm}

Two lines in the algorithm deserve some more discussion. In line~5, we compute the minimum of the calibration measure without solving an optimization problem. This is because we know the ground-truth probability distribution in this case. In the limit, when the sample size grows to infinity, the calibration measure $g$ will even be zero, so in fact we are looking for the natural deviation from zero for a sample of size $N$. 

Conversely, in line~7, we have to solve an optimization problem to find the minimum over $\boldsymbol{\lambda} \in \Delta_M$. Specific solvers are needed here, because the objective functions are in most cases not differentiable, e.g., for the measures in (\ref{eq:HL}), (\ref{eq:ECEcwise}) and (\ref{eq:ECEconf}) this is not the case. The objective functions constructed from (\ref{eq:SKCEuq}) and (\ref{eq:SKCEul}) are differentiable, but not convex. In this work we use constrained optimization by linear approximation (COBYLA), which is a numerical optimization method for constrained problems where the derivative of the objective function is not known~\citep{Powell94COBYLA}. In principle, any constrained derivative-free optimization algorithm can be used in line 7, and exploring other solvers will be considered in future work.

\section{EXPERIMENTS}

Two types of experiments are considered in this work. First, we analyse the empirical Type I and II error of our statistical test for synthetic datasets. Second, we apply our test on several real-world datasets and popular ensemble-based methods to investigate whether these methods return calibrated representations. Detailed information regarding the experimental setup and datasets can be found in the supplementary materials (see Section~1).

\subsection{Type I And II Error Analysis}

In a first set of experiments, we analyse the empirical Type I and II error rate for Alg.~\ref{alg:gcr}, with number of bootstrap samples $D=100$, where we consider the calibration measures that have been discussed in Section~\ref{sec:aleatoric}: $SKCE_{ul}$, $HL_{cwise}$, $ECE_{conf}$ and $ECE_{cwise}$. For computational reasons, we choose not to incorporate $SKCE_{uq}$. For simplicity, we use a bin size of $B=5$ and $B=10$ for $HL_{cwise}$ and $ECE_{conf,cwise}$. Similar as in \citet{Widmann2019}, for $SKCE_{ul}$, we use the matrix-valued kernel $\Gamma(\boldsymbol{p},\boldsymbol{p}')=\exp(-\|\boldsymbol{p}-\boldsymbol{p}'\|/2)\boldsymbol{I}_{K}$, combining the commonly-used total variation distance and the $K \times K$ identity matrix $\boldsymbol{I}_{K}$. Moreover, we analyse the statistical Type I and II error under three different scenarios, which we denote by S1, S2 and S3, respectively. S1 generates data under the null hypothesis that the probabilistic classifier set is calibrated, whereas S2 and S3 generate data under the alternative hypothesis that the probabilistic classifier set is not calibrated. To this end, we construct $R=1000$ synthetic datasets that contain $N=100$ instances. For each instance, we generate predictions for $M$ probabilistic models and a ground-truth label: $\{(\boldsymbol{p}^{(1)}(\boldsymbol{X}_{i}),\ldots,\boldsymbol{p}^{(M)}(\boldsymbol{X}_{i}),Y_{i})\}_{i=1}^{N}$, with ensemble size $M=10$, for $K=10$ classes. For each dataset, we assume a mean $\boldsymbol{p}_{e}|\boldsymbol{X}_{i} \sim Dir(\boldsymbol{a}_{e})$, with $a_{ek}=1/K$ for all $k=1,\ldots,K$, and sample an ensemble from the prior $\boldsymbol{p}^{(1)},\ldots,\boldsymbol{p}^{(M)}|\boldsymbol{X}_{i}\sim Dir(K\boldsymbol{p}_{e}/u)$. By means of this prior, we are able to control the center $\boldsymbol{p}_{e}$ and uncertainty (or spread) $u$ of the convex sets, similarly as in \citep{Sensoy18EDL}. Then, we simulate the corresponding labels $Y_{i}$ conditionally on the probabilistic classifier set in three ways:
\begin{itemize}[noitemsep,topsep=0pt,leftmargin=6.5mm]
    \item[S1:] The null hypothesis is true. For each dataset, we sample uniformly at random $\boldsymbol{\lambda}\in \Delta_M$, and select $\boldsymbol{p}_{\boldsymbol{\lambda}}$ as the ground-truth probability distribution, i.e., the labels are generated according to  $Cat(\boldsymbol{p}_{\boldsymbol{\lambda}})$.
    \item[S2:] The null hypothesis is false. For each dataset, we generate labels using a categorical distribution that is randomly chosen outside the convex set on the line segment between the closest corner in $\Delta^{K}$ and $\boldsymbol{p}_{e}$. 
    \item[S3:] The null hypothesis is false, too. Similarly as in S2, a categorical distribution is randomly chosen outside the convex set, but this time on the line segment between a randomly chosen corner in $\Delta^{K}$ and $\boldsymbol{p}_{e}$. In this case, it should be somewhat easier to reject the null hypothesis than in scenario S2. 
\end{itemize}

\begin{figure*}
\centering
\begin{subfigure}{.2\textwidth}
  \centering
  \includegraphics[width=\textwidth]{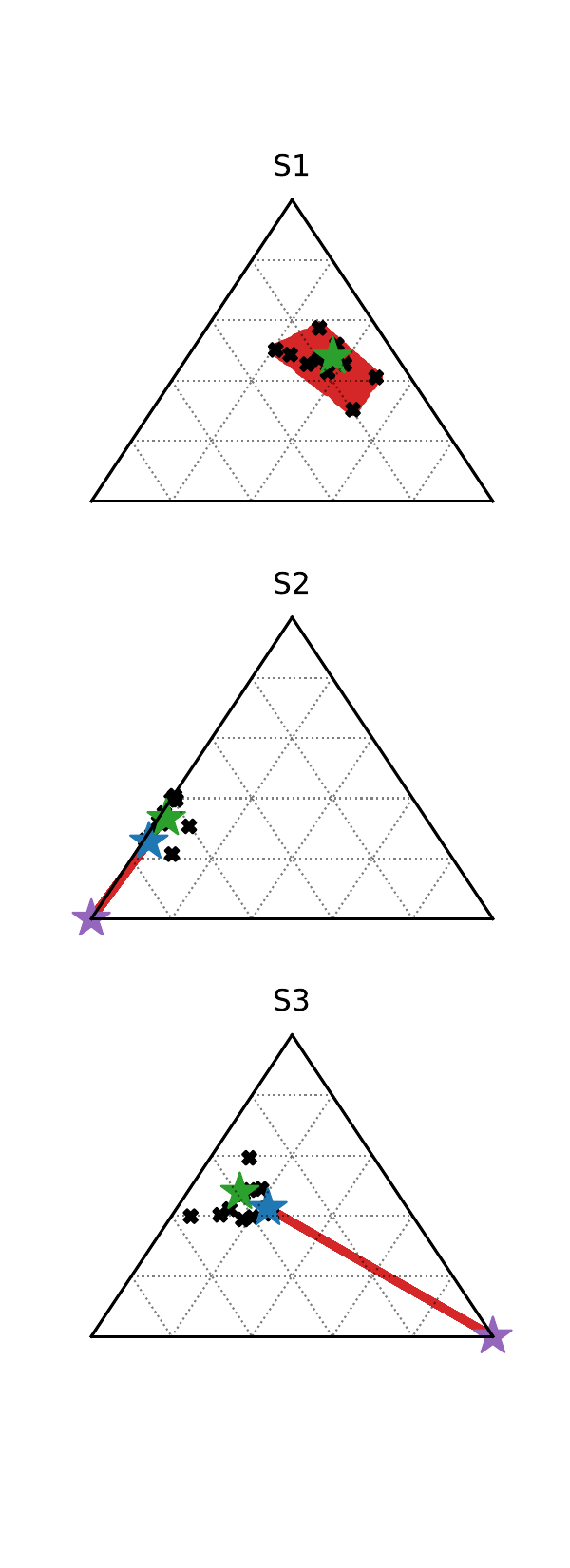}
\end{subfigure}
\begin{subfigure}{.5\textwidth}
  \centering
  \includegraphics[width=\textwidth]{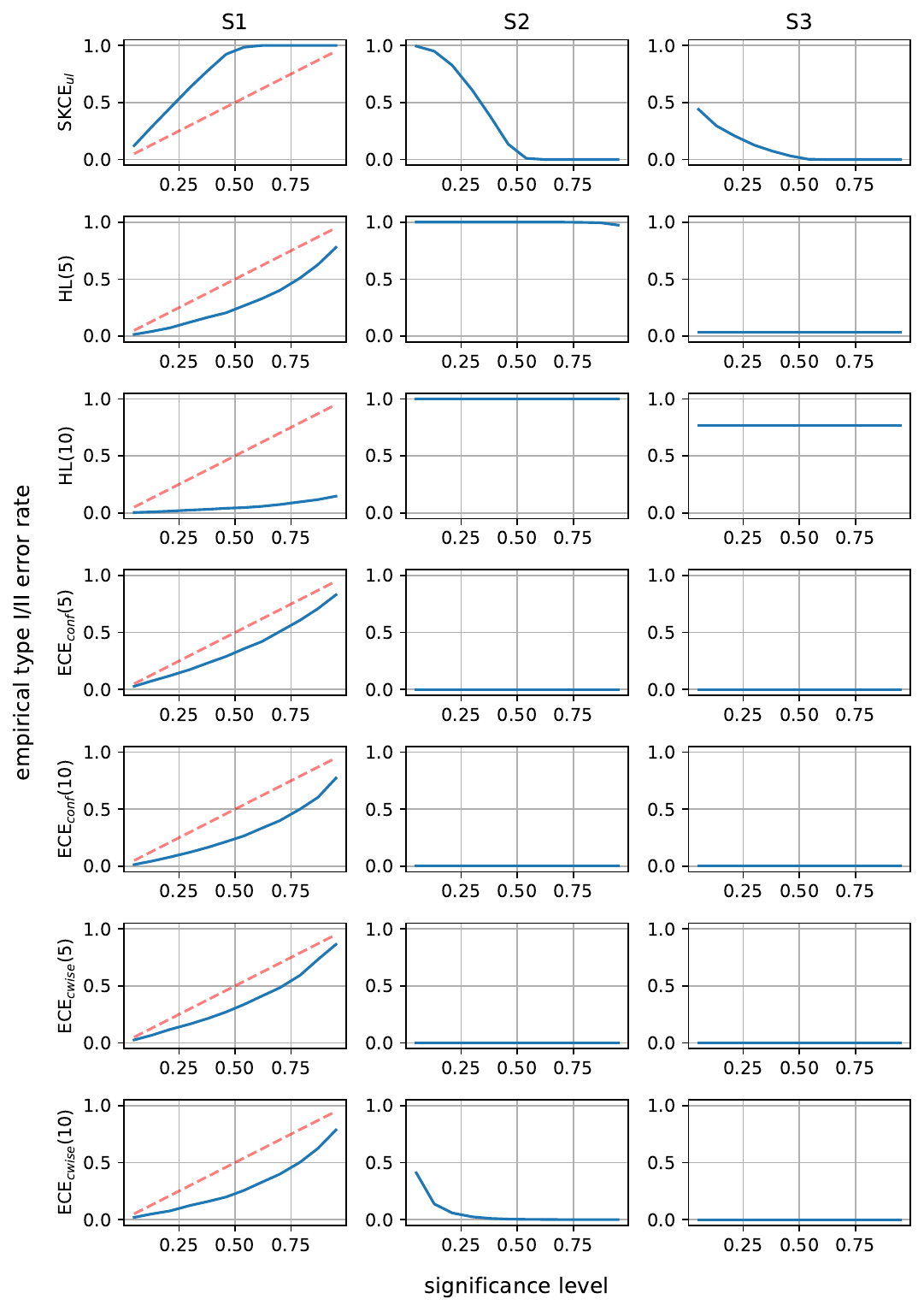}
\end{subfigure}%
\caption{Left: A visualization of the setup in scenarios S1, S2 and S3. In all three scenarios, black stars correspond to an ensemble that has been sampled as outlined in the main text. In S1, the red dots correspond to a convex set from which the ground-truth distribution is uniformly sampled. In S2 and S3, the ground-truth distribution is uniformly sampled from the red line segment, outside the convex set. For S2, the line segment connects the calculated boundary (blue star) and closest corner in the simplex (purple star) of the convex set. For S3, the line segments connects a boundary of the convex set and a random corner in the simplex. Right: empirical Type I (S1) and Type II (S2, S3) error rate for different calibration measures in function of the significance level for $R=1000$ randomly sampled datasets from S1, S2 and S3 and with $N=100, M=10, K=10$ and $u=0.01$. For all tests we use $D=100$ bootstrap samples. }
\label{fig:t1t2:alpha}
\end{figure*}

 The three different scenarios are illustrated in Fig.~\ref{fig:t1t2:alpha} (left) for $K=3$ and $M=10$. For scenarios S2 and S3, an additional algorithm is needed to compute the line segment outside the convex set. This algorithm is explained in the supplementary materials (see Section~1).

\begin{table*}[t!]
      \centering
      \vskip 0.1in
    \caption{Results obtained for four different classifiers: single classifier (S), dropout network with rate 0.1 (DN(0.1)) and 0.6 (DN(0.6)) and a deep ensemble (DE). Classifiers are tested on six benchmark datasets. For the image datasets, we use two different architectures: VGG16 (VGG) and MobileNetV2 (MOB). For each classifier we consider the average and two different weighted averages, defined by the convex combination found in line 7 of Alg.~\ref{alg:gcr}, for $ECE_{conf}$ and $ECE_{cwse}$, respectively. We report the average accuracy (Acc.) and calibration measure ($ECE_{conf,cwise}$) obtained on the test sets. For the two weighted averages, which correspond to confidence and classwise calibration, respectively, we also present the outcome (ie. reject or not/$\neg$ reject $H_0$) of our test on a separate calibration dataset.}
    \vspace{0.1in}
    \label{tab:gen}
      \resizebox{0.9\textwidth}{!}{%
      \begin{tabular}{ccc|ccc|rcc|rcc}
      \toprule
    &  &  & \multicolumn{3}{c|}{\textsc{Avg.}} & \multicolumn{6}{c}{\textsc{Weighted avg.}} \\
    &  &  & \multicolumn{3}{c|}{} & \multicolumn{3}{c|}{$\boldsymbol{\lambda}_{conf}$} & \multicolumn{3}{c}{$\boldsymbol{\lambda}_{cwise}$} \\
      \textsc{Dataset} & \textsc{Arch.} & \textsc{Classifier} & \textsc{Acc.} & $ECE_{conf}$ & $ECE_{cwise}$ & $H_{0}$ & \textsc{Acc.} & $ECE_{conf}$ & $H_{0}$ & \textsc{Acc.} & $ECE_{cwise}$ \\
      \midrule
      \multirow{8}{*}{\textsc{CIFAR-10}} & \multirow{4}{*}{MOB} & \textsc{S} & 0.7804 & 0.0194 & 0.0071 & \textsc{Rej.}  & 0.7804 & 0.0194 & \textsc{Rej.}  & 0.7804 & 0.0071 \\
      & & \textsc{DN(0.1)} & 0.7740 & 0.0612 & 0.0134 & \textsc{Rej.}  & 0.7726 & 0.0629 & \textsc{Rej.}  & 0.7730 & 0.0137 \\
      & & \textsc{DN(0.6)} & 0.3039 & 0.0196 & 0.0108 & $\neg$\textsc{Rej.}  & 0.3025 & 0.0142 & \textsc{Rej.}  & 0.3007 & 0.0111 \\
      & & \textsc{DE} & 0.8329 & 0.1680 & 0.0327 & $\neg$\textsc{Rej.}  & 0.7642 & 0.0150 & \textsc{Rej.}  & 0.7668 & 0.0062 \\
      \cmidrule{3-12}
      & \multirow{4}{*}{VGG} & \textsc{S} & 0.8438 & 0.0828 & 0.0183 & \textsc{Rej.}  & 0.8438 & 0.0828 & \textsc{Rej.}  & 0.8438 & 0.0183 \\
      & & \textsc{DN(0.1)} & 0.8476 & 0.0821 & 0.0183 & \textsc{Rej.}  & 0.8474 & 0.0832 & \textsc{Rej.}  & 0.8462 & 0.0181 \\
      & & \textsc{DN(0.6)} & 0.8355 & 0.0950 & 0.0202 & \textsc{Rej.}  & 0.8353 & 0.0954 & \textsc{Rej.}  & 0.8351 & 0.0203 \\
      & & \textsc{DE} & 0.8754 & 0.0044 & 0.0042 & $\neg$\textsc{Rej.}  & 0.8756  & 0.0052 & $\neg$\textsc{Rej.}  & 0.8766 & 0.0044 \\
      \midrule
      \multirow{8}{*}{\textsc{Caltech-101}} & \multirow{4}{*}{MOB} & \textsc{S} & 0.9477 & 0.0135 & 0.0010 & \textsc{Rej.}  & 0.9477 & 0.0135 & \textsc{Rej.}  & 0.9477 & 0.0010 \\
      & & \textsc{DN(0.1)} & 0.9384 & 0.0081 & 0.0010 & $\neg$\textsc{Rej.}  &  0.9389 & 0.0087 & \textsc{Rej.}  & 0.9361 & 0.0010 \\
      & & \textsc{DN(0.6)} & 0.9338 & 0.0129 & 0.0012 & $\neg$\textsc{Rej.}  & 0.9338 & 0.0107 & \textsc{Rej.}  & 0.9324 & 0.0012 \\
      & & \textsc{DE} & 0.9625 & 0.0175 & 0.0009 & $\neg$\textsc{Rej.}  & 0.9509 & 0.0076 & $\neg$\textsc{Rej.}  & 0.9579 & 0.0009 \\
      \cmidrule{3-12}
      & \multirow{4}{*}{VGG} & \textsc{S} & 0.9287 & 0.0395 & 0.0013 & \textsc{Rej.}  & 0.9287 & 0.0395 & \textsc{Rej.}  & 0.9287 & 0.0013 \\
      & & \textsc{DN(0.1)} & 0.9218 & 0.0192 & 0.0013 & $\neg$\textsc{Rej.}  & 0.9204 & 0.0212 & \textsc{Rej.}  & 0.9218 & 0.0014 \\
      & & \textsc{DN(0.6)} & 0.9259 & 0.0155 & 0.0015 & $\neg$\textsc{Rej.}  & 0.9162 & 0.0062 & \textsc{Rej.}  & 0.9157 & 0.0015 \\
      & & \textsc{DE} & 0.9500 & 0.0231 & 0.0011 & $\neg$\textsc{Rej.}  & 0.9329 & 0.0079 & $\neg$\textsc{Rej.}  & 0.9306 & 0.0011 \\
      \midrule
      \multirow{8}{*}{\textsc{Caltech-256}} & \multirow{4}{*}{MOB} & \textsc{S} & 0.7829 & 0.0638 & 0.0009 & \textsc{Rej.}  & 0.7829 & 0.0638 & \textsc{Rej.}  &  0.7829 & 0.0009 \\
      & & \textsc{DN(0.1)} & 0.7820 & 0.0473 & 0.0008 & \textsc{Rej.}  & 0.7816 & 0.0480 & \textsc{Rej.}  & 0.7823 & 0.0008 \\
      & & \textsc{DN(0.6)} & 0.7395 & 0.0201 & 0.0009 & $\neg$\textsc{Rej.}  & 0.7313 & 0.0085 & \textsc{Rej.}  & 0.7336 & 0.0009 \\
      & & \textsc{DE} & 0.8383 & 0.0358 & 0.0007 & $\neg$\textsc{Rej.}  & 0.8082 & 0.0130 & \textsc{Rej.}  & 0.8312 & 0.0006 \\
      \cmidrule{3-12}
      & \multirow{4}{*}{VGG} & \textsc{S} & 0.7552 & 0.0948 & 0.0011 & \textsc{Rej.}  & 0.7552 & 0.0948 & \textsc{Rej.}  & 0.7552 & 0.0011 \\
      & & \textsc{DN(0.1)} & 0.7458 & 0.0645 & 0.0011 & \textsc{Rej.}  & 0.7452 & 0.0663 & \textsc{Rej.}  & 0.7421 & 0.0011 \\
      & & \textsc{DN(0.6)} & 0.7427 & 0.0106 & 0.0010 & $\neg$\textsc{Rej.}  & 0.7414 & 0.0103 & \textsc{Rej.}  & 0.7421 & 0.0010 \\
      & & \textsc{DE} & 0.8128 & 0.0326 & 0.0007 & $\neg$\textsc{Rej.}  & 0.7864 & 0.0104 & \textsc{Rej.}  & 0.8108 & 0.0007 \\
      \midrule
      \multirow{8}{*}{\textsc{PlantCLEF2015}} & \multirow{4}{*}{MOB} & \textsc{S} & 0.4842 & 0.1300 & 0.0005 & \textsc{Rej.}  & 0.4842 & 0.1300 & \textsc{Rej.}  & 0.4842 & 0.0005 \\
      & & \textsc{DN(0.1)} & 0.4357 & 0.0866 & 0.0005 & $\neg$\textsc{Rej.}  & 0.4320 & 0.0920 & \textsc{Rej.}  & 0.4319 & 0.0005 \\
      & & \textsc{DN(0.6)} & 0.4382 & 0.0199 & 0.0005 & $\neg$\textsc{Rej.}  & 0.4343 & 0.0245 & \textsc{Rej.}  & 0.4348 & 0.0005 \\
      & & \textsc{DE} & 0.5973 & 0.0993 & 0.0004 & $\neg$\textsc{Rej.}  & 0.5364 & 0.0171 & \textsc{Rej.}  & 0.5757 & 0.0004  \\
      \cmidrule{3-12}
      & \multirow{4}{*}{VGG} & \textsc{S} & 0.4085 & 0.1362 & 0.0006 & \textsc{Rej.}  & 0.4085 & 0.1362 & \textsc{Rej.}  & 0.4085 & 0.0006 \\
      & & \textsc{DN(0.1)} & 0.4156 & 0.1009 & 0.0006 & $\neg$\textsc{Rej.}  & 0.4151 & 0.1020 & \textsc{Rej.}  & 0.4146 & 0.0006 \\
      & & \textsc{DN(0.6)} & 0.3927 & 0.0566 & 0.0005 & $\neg$\textsc{Rej.}  & 0.3928 & 0.0566 & \textsc{Rej.}  & 0.3894 & 0.0005 \\
      & & \textsc{DE} & 0.5488 & 0.1045 & 0.0004 & $\neg$\textsc{Rej.}  & 0.4705 & 0.0197 & \textsc{Rej.}  & 0.5236 & 0.0004 \\
      \midrule
      \multirow{4}{*}{\textsc{Bacteria}} & \multirow{4}{*}{--} & \textsc{S} & 0.8785 & 0.0507 & 0.0002 & $\neg$\textsc{Rej.}  & 0.8785 & 0.0507 & $\neg$\textsc{Rej.}  & 0.8785 & 0.0002 \\
      & & \textsc{DN(0.1)} & 0.8398 & 0.1504 & 0.0002 & \textsc{Rej.}  & 0.8345 & 0.1508 & \textsc{Rej.}  & 0.8371 & 0.0002 \\
      & & \textsc{DN(0.6)} & 0.8407 & 0.2085 & 0.0002 & $\neg$\textsc{Rej.}  & 0.7491 & 0.0829 & $\neg$\textsc{Rej.}  & 0.7879 & 0.0002 \\
      & & \textsc{DE} & 0.8926 & 0.1437 & 0.0002 & $\neg$\textsc{Rej.}  & 0.8565 & 0.0642 & $\neg$\textsc{Rej.}  & 0.9032 & 0.0001 \\
      \midrule
      \multirow{4}{*}{\textsc{Proteins}} & \multirow{5}{*}{--} & \textsc{S} & 0.8001 & 0.0454 & 0.0001 & \textsc{Rej.}  & 0.8001 & 0.0454 & \textsc{Rej.}  & 0.8001 & 0.0001  \\
      & & \textsc{DN(0.1)} & 0.7909 & 0.0788 & 0.0001 & \textsc{Rej.}  & 0.7895 & 0.0779 & \textsc{Rej.}  & 0.7915 & 0.0001 \\
      & & \textsc{DN(0.6)} & 0.8117 & 0.0393 & 0.0001 & \textsc{Rej.}  & 0.8033 & 0.0353 & \textsc{Rej.}  & 0.8119 & 0.0001  \\
      & & \textsc{DE} & 0.8076 & 0.0764 & 0.0001 & \textsc{Rej.}  & 0.7968 & 0.0528 & \textsc{Rej.}  & 0.8050 & 0.0001 \\
      \bottomrule
      \end{tabular}
      }
      \vskip -0.13in
\end{table*}

The results are shown in Fig.~\ref{fig:t1t2:alpha} (right). Our test with $SKCE_{ul}$ is not correct, because it makes more errors than defined by the significance level. However, for the other measures, our test is more reliable, because the Type I error is mostly lower than the significance level.  $HL_{cwise}$ does not seem to yield high power for S2, for different bin sizes. Empirically, our test with $ECE_{conf,cwise}$ results in reliable tests when it comes to both the Type I and Type II error. In the supplementary materials (see Section~2), we also show some results obtained for $u=0.1$ (Fig.~1), $M=100$ (Fig.~2) and $K=100$ (Fig.~3). Similar findings are obtained w.r.t.\ the calibration measure $SKCE_{ul}$. For most cases, larger probabilistic classifier sets result in more conservative tests, as can be observed in Fig.~1. Increasing the ensemble size, however, does not seem to have a significant influence on the results, following Fig.~2. Finally, when looking at Fig.~3, by increasing the number of classes, the empirical test error approaches the significance level, making most tests less conservative.

\subsection{Calibration Of Probabilistic Classifier Sets Based On Deep Neural Networks}

In a last set of experiments, we apply our test in Alg.~\ref{alg:gcr} on six benchmark datasets, where we consider a single classifier (S) and three ensemble-based models: two dropout networks with rate 0.1 and 0.6 (DN(0.1), DN(0.6)) and a deep ensemble (DE), where we use an ensemble size of ten. For the deep ensembles, a probabilistic classifier set is obtained by training ten different models, i.e., with different initializations of the weights~\citep{Lakshminarayanan17}. For the dropout networks, a probabilistic classifier set is obtained by using dropout in the last layer and sampling ten predictions~\citep{Gal2016}.  For the image datasets, we consider a small and large neural network architecture that are commonly used in the literature: MobileNetV2 (MOB) with $3.4\times 10^{6}$ parameters and VGG16 (VGG) with $138 \times 10^{6}$ parameters, respectively~\citep{Sandler18mobilenetv2,Simonyan14VGG16}. For the last two biological datasets, a simple neural network with one hidden layer is considered, together with textual feature representations. For more information related to the training of the models and datasets, we refer the reader to the supplementary materials (see Section~1). Furthermore, we analyse the calibration of the probabilistic classifier sets by means of the $ECE_{conf}$ and $ECE_{cwise}$ calibration measures, since those measures gave reliable tests in terms of the Type I and Type II error in the previous simulations. 

The results are shown in Table~\ref{tab:gen}. For each classifier, we report results for the ensemble average, i.e., one the average of the ensemble predictions, and the weighted ensemble average, with the weights $\boldsymbol{\lambda}$ obtained in line 7 of Alg.~\ref{alg:gcr}. More precisely, two different weighted ensemble averages are considered by running our test with the $ECE_{conf}$ and $ECE_{cwise}$ measure, respectively, on a specific calibration set. For each weighted average, we show the outcome of our test w.r.t.\ the set of hypotheses in (\ref{def_test_single_prob}), together with the average accuracy and calibration measure obtained on the test set. For the single classifier case, our test in Alg.~\ref{alg:gcr} corresponds to the specific test from \citet{Vaicenavicius2019}. 

The single classifiers based on deep neural networks are in general not calibrated, and this corresponds to similar findings that have been reported in the literature~\citep{Guo2017}. For most cases, the classifiers are not calibrated in terms of $ECE_{cwise}$, which is not so surprising, because classwise calibration is already a very strong form of calibration. Our findings are in line with what has been reported in the literature on linear opinion pooling and deep ensembles, namely, a linear opinion pool is often not calibrated in the strong sense, even in the ideal case in which the individual models are calibrated \citep{Hora04LP,Ranjan10LP,Lichtendahl13LP,Rahaman21UQDE,Schulz22LPDL}. When it comes to $ECE_{conf}$ and dropout networks, in some cases, our test does not reject the null hypothesis when a higher dropout rate is considered. This indicates that dropout networks tend to be better confidence calibrated when using a higher dropout rate, which is also confirmed by comparing the average $ECE_{conf}$ on the test set between DN(0.1) and DN(0.6). A possible explanation could be given by the fact that the prior, from which models are sampled in the dropout network with higher dropout rate, is more diverse and results in large probabilistic classifier sets that include the ground-truth distribution. Finally, when it comes to $ECE_{conf}$, deep ensembles appear to be calibrated for most datasets and architectures, since the null hypothesis is almost never rejected. For those classifiers, there is also a significant difference in terms of $ECE_{conf}$ between the average and weighted average, which indicates that using a simple averaging strategy results in less calibrated deep ensembles. 

\section{DISCUSSION}
In this paper we addressed the following question: What does it mean that a probabilistic classisfier set represents epistemic uncertainty in a faithful manner? To answer this question, we referred to the notion of calibration of probabilistic classifiers and extended it to probabilistic classifier sets. We called a probabilistic classifier set $S(\mathcal{P})$ calibrated if the set contained at least one calibrated classifier. To verify this property for the important case of ensemble-based models, we proposed a novel nonparametric calibration test that generalizes existing tests for probabilistic classifiers to the case of probabilistic classifier sets. In our experiments, we analyzed the Type I and II error of the newly-proposed test for different scenarios. The best results were obtained when using expected calibration error as underlying calibration measure, but for most measures the Type I and II error were both sufficiently low.  Making use of this test, we empirically show that probabilistic classifier sets based on deep neural networks are often not calibrated. 


\subsubsection*{Acknowledgements}

For this work W.W. received funding from the Flemish government under the ``Onderzoeksprogramma Artifici\"ele Intelligentie (AI) Vlaanderen'' Programme (Number 174L00121).

\bibliography{refs}

\appendix
\onecolumn

\section{EXPERIMENTAL SETUP}
\label{sec:app:expsetup}
\subsection{Type I And II Error Analysis}
\label{sec:app:expsetup:t1t2}
Here we explain in detail how the ground-truth distribution is generated in scenarios S2 and S3. In both cases, the null hypothesis is false, so one needs to sample a ground-truth distribution outside the convex set. To find a distribution outside the convex set, for every $\boldsymbol{x}$, we need to find the largest $\lambda_{b}\in [0,1]$, such that $\boldsymbol{p}_{\lambda_{b}} \in S(\mathcal{P})$, with $\boldsymbol{p}_{\lambda_{b}}=(1-\lambda_{b})\boldsymbol{p}_{e}+\lambda_{b}\boldsymbol{p}_{c}$, $\boldsymbol{p}_{e}$ the mean and $\boldsymbol{p}_{c}$ the randomly chosen corner. This can be calculated by means of an exhaustive line search for $\lambda$ over the interval $[0,1]$ and a linear program. Pseudocode for this procedure is given by Alg.~\ref{alg:inconvexset}. After finding the boundary, one can simply sample a random distribution on the line segment between $\boldsymbol{p}_{\lambda_{b}}$ and $\boldsymbol{p}_{c}$.
\begin{algorithm}[ht]
\begin{small}
    \caption{findBoundary -- \textbf{input:} $S(\mathcal{P})$, $\boldsymbol{p}_{0}$, $\boldsymbol{p}_{c}$, LP}
\begin{algorithmic}[1] 
\State $\boldsymbol{P} \leftarrow [\boldsymbol{p}^{(1)};\ldots;\boldsymbol{p}^{(M)}]$, i.e., $\boldsymbol{P}\in[0,1]^{M\times K}$ \Comment{$\boldsymbol{P}$ represents $\mathcal{P}$ in matrix-notation} 
\State $\boldsymbol{A} = [\boldsymbol{P}^{T};\mathbf{1}_{M}]$ with row vector $\mathbf{1}_{M}=[1,\ldots,1]$
\State $\boldsymbol{p}_{\lambda_{b}} \leftarrow \boldsymbol{p}_{0}$
\For{$\lambda'=0$ to $1$} \Comment{Begin exhaustive line search}
    \State $\boldsymbol{p}_{\lambda'} \leftarrow (1-\lambda')\boldsymbol{p}_{0}+\lambda'\boldsymbol{p}_{c}$
    \State $\boldsymbol{z} = [\boldsymbol{p}_{\lambda'}; 1]$
    \If{LP$(\boldsymbol{A},\boldsymbol{z})$ finds a solution} \Comment{Check if $\boldsymbol{p}_{\lambda'}$ falls inside the convex}
        \State $\boldsymbol{p}_{\lambda_{b}} \leftarrow \boldsymbol{p}_{\lambda'}$
    \Else
        \State \textbf{break} \Comment{We are outside the convex set, hence, break and return previous solution}
    \EndIf
\EndFor
\State \textbf{return} $\boldsymbol{p}_{\lambda_{b}}$
\end{algorithmic}
\label{alg:inconvexset}
\end{small}
\end{algorithm}
\subsection{Calibration of probabilistic classifier sets based on deep neural networks}
\label{sec:app:expsetup:data}

\begin{table}[ht]
\centering
\vskip 0.1in
\caption{\small Overview of of image (top) and text (bottom) datasets used in the experiments. Notation: $K$ -- number of classes, $D$ -- number of features, $N$ -- number of samples for training, calibration and test set.}
\label{tab:exp:datasets}
\resizebox{0.7\textwidth}{!}{%
\begin{tabular}{lrrrrr}
\toprule
    \textbf{Dataset} & $\mathbf{K}$ & $\mathbf{D}$ & $\mathbf{N_{train}}$ & $\mathbf{N_{cal.}}$ & $\mathbf{N_{test}}$ \\
    \midrule
    \textbf{CIFAR-10}~\citep{Krizhevsky10cifar10} & 10 & 1000 & 50000 & 4992 & 4992\\
    \textbf{Caltech-101}~\citep{Li03Caltech101} & 97 & 1000 & 4338 & 2160 & 2160 \\
    \textbf{Caltech-256}~\citep{Griffin07Caltech256} & 256 & 1000 & 14890 & 7440 & 7440 \\
    \textbf{PlantCLEF2015}~\citep{Goeau15PlantClef} & 1000 & 1000 & 91758 & 10720 & 10720 \\
    \midrule
    \textbf{Bacteria}~\citep{RIKEN13Bacteria} & 2659 & 1000 & 10587 & 1136 & 1136\\
    \textbf{Proteins}~\citep{Li18DEEPre} & 3485 & 1000 & 11830 & 5088 & 5088\\
    \bottomrule%
\end{tabular}%
}
\end{table}

We use a MobileNetV2 or VGG16 convolutional neural network~\citep{Sandler18mobilenetv2,Simonyan14VGG16}, pretrained on ImageNet~\citep{Deng09ImageNet}, in order to obtain hidden representations for all image datasets. For the bacteria dataset, tf-idf representations are obtained by means of extracting 3-, 4- and 5-grams from the 16S rRNA sequences that were provided in the dataset~\citep{Fiannaca18Bacteria}. For the proteins dataset, tf-idf representations are obtained by considering 3-grams only. Furthermore, to comply with literature, the tf-idf representations are concatenated with functional domain encodings, which contain distinct functional and evolutional information about the protein sequence~\citep{Li18DEEPre}. Next, obtained feature representations for the biological datasets are passed through a single-layer neural net with 1000 output neurons and a ReLU activation function. We use the categorical cross-entropy loss by means of stochastic gradient descent with momentum, where the learning rate and momentum are set to $1e-5$ and 0.99, respectively. We set the number of epochs to 2 and 20, for the Caltech and other datasets, respectively. We train all models end-to-end on a GPU, by using the PyTorch library~\citep{Paszke17pytorch} and infrastructure with the following specifications:
\begin{itemize}
    \item \textbf{CPU:} i7-6800K 3.4 GHz (3.8 GHz Turbo Boost) – 6 cores / 12 threads,
    \item \textbf{GPU:} 2x Nvidia GTX 1080 Ti 11GB + 1x Nvidia Tesla K40c 11GB,
    \item \textbf{RAM:} 64GB DDR4-2666.
\end{itemize}

\section{ADDITIONAL EXPERIMENTS}
\label{sec:app:addexp}
\begin{figure}[h!]
\centering
  \includegraphics[width=0.7\textwidth]{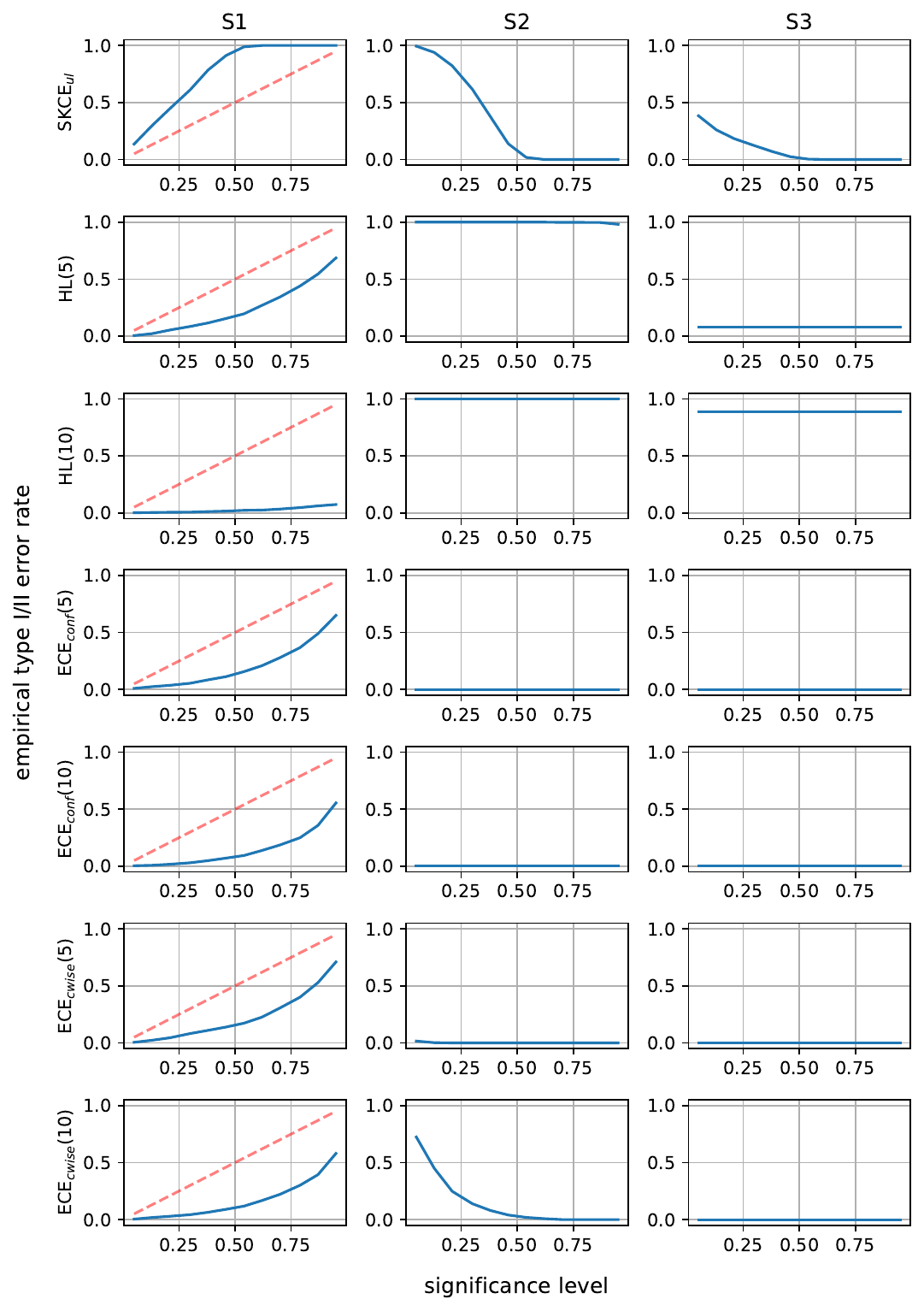}
  \caption{Empirical Type I (S1) and Type II (S2, S3) error for different calibration measures in function of the significance level for $R=1000$ randomly sampled datasets from S1, S2 and S3 and with $N=100, M=10, K=10$ and $u=0.1$. For all tests we use $D=100$ bootstrap samples.}
  \label{fig:t1t2:alpha:ubig}
\end{figure}

\begin{figure}[h!]
\centering
  \includegraphics[width=0.7\textwidth]{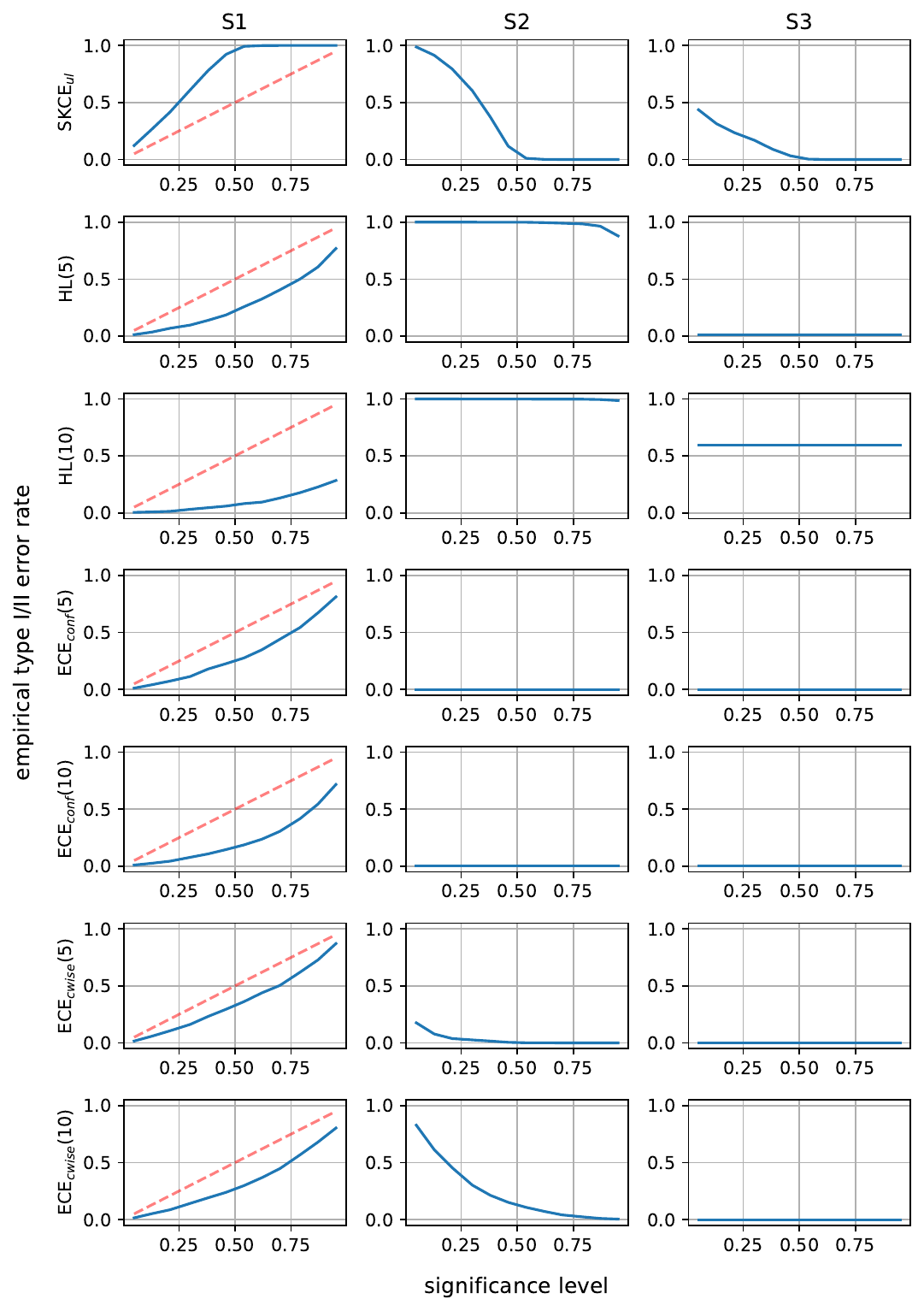}
  \caption{Empirical Type I (S1) and Type II (S2, S3) error for different calibration measures in function of the significance level for $R=1000$ randomly sampled datasets from S1, S2 and S3 and with $N=100, M=100, K=10$ and $u=0.01$. For all tests we use $D=100$ bootstrap samples.}
  \label{fig:t1t2:alpha:mbig}
\end{figure}

\begin{figure}[h!]
\centering
  \includegraphics[width=0.7\textwidth]{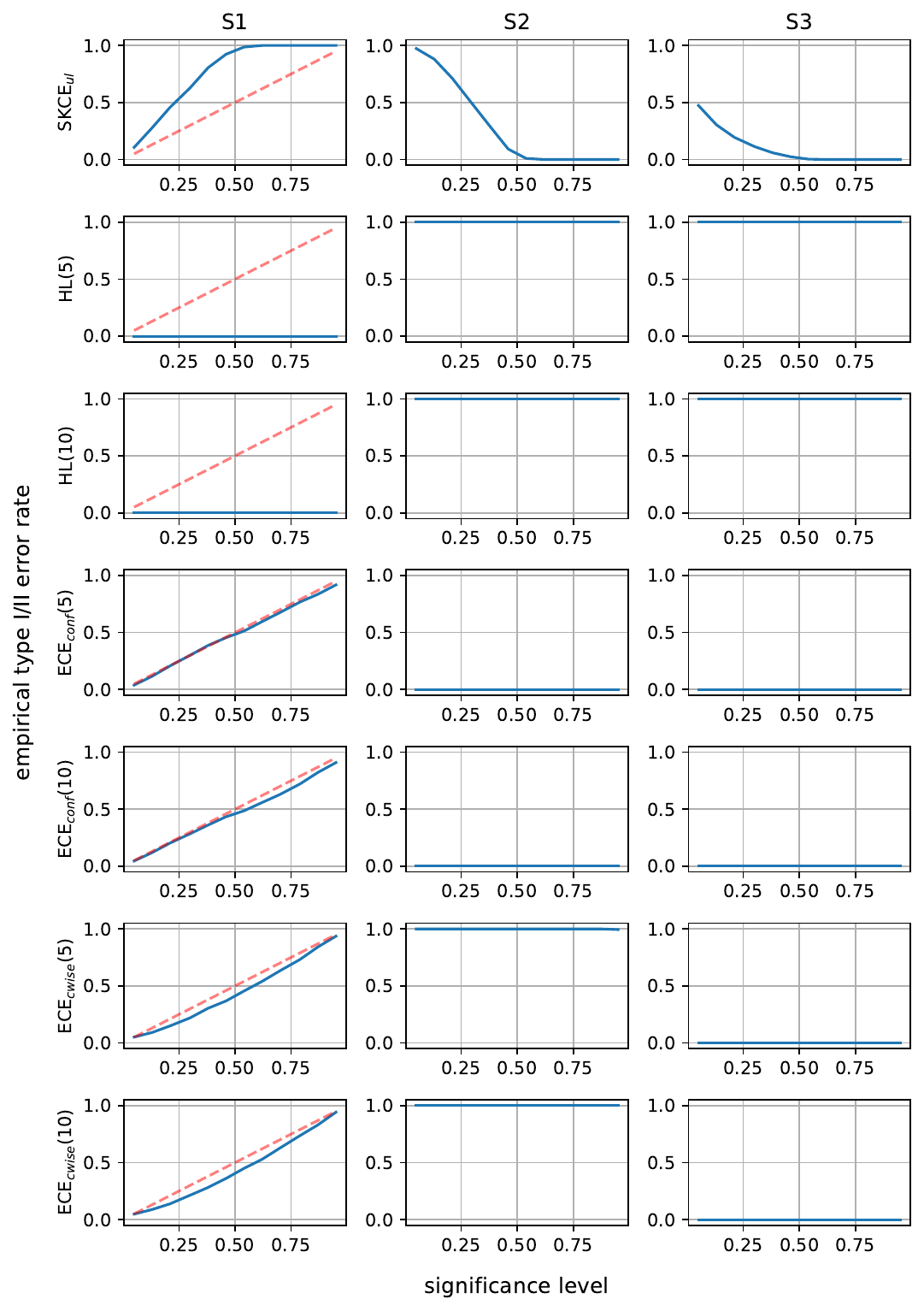}
  \caption{Empirical Type I (S1) and Type II (S2, S3) error for different calibration measures in function of the significance level for $R=1000$ randomly sampled datasets from S1, S2 and S3 and with $N=100, M=10, K=100$ and $u=0.01$. For all tests we use $D=100$ bootstrap samples.}
  \label{fig:t1t2:alpha:kbig}
\end{figure}

\end{document}